\pdfoutput=1

\documentclass[11pt]{article}
\usepackage[dvipsnames]{xcolor}
\usepackage{naacl2021}

\usepackage{times}
\usepackage{latexsym}
\usepackage{amsmath}
\usepackage{multirow}
\usepackage{comment}
\DeclareMathOperator{\avg}{avg}
\DeclareMathOperator*{\argmax}{argmax}
\usepackage{hyperref}
\usepackage[T1]{fontenc}
\usepackage{graphicx}

\usepackage[utf8]{inputenc}

\usepackage{microtype}

\title{ Predicting Discourse Trees from 
Transformer-based Neural Summarizers}

\author{Wen Xiao, Patrick Huber, Giuseppe Carenini\\
  Department of Computer Science \\
  University of British Columbia \\
  Vancouver, BC, Canada, V6T 1Z4 \\
  {\tt \{xiaowen3, huberpat, carenini\}@cs.ubc.ca}}

\date{}

\begin{document}
\maketitle
\begin{abstract}
Previous work indicates that discourse information benefits summarization. In this paper, we explore whether this synergy between discourse  and  summarization  is  bidirectional, by 
inferring document-level discourse trees 
from pre-trained neural summarizers. In particular, we generate unlabeled RST-style discourse trees from 
the self-attention matrices of the  
transformer 
model. Experiments across models and datasets reveal
that the summarizer learns both, dependency- and 
constituency-style discourse information, 
which is typically encoded 
in a single head, 
covering long- and short-distance discourse dependencies. 
Overall, the experimental results
suggest that the learned discourse information is general and transferable inter-domain\footnote{The code can be found in \url{https://github.com/Wendy-Xiao/summ_guided_disco_parser}}.

\end{abstract}
\section{Introduction}

Extractive summarization is a common and important task within the area of Natural Language Processing (NLP)
, which can be useful in a multitude of diverse real-life scenarios. 
Current extractive summarizers typically 
use exclusively neural approaches, in which the importance of extracted units (i.e., sentences or clauses) and relationship between them are learned by the model 
from a large amount of data (e.g., \citet{bertsum}). 

Inspired by previous work in pre-neural times, indicating that discourse information, especially discourse trees according to the  Rhetorical Structure Theory (RST) \cite{rst}, 
can benefit  the summarization task \cite{Marcu1999}, several very recent neural summarizers have tried to explicitly encode discourse information to support summarization. 
Overall, it seems that adding these encodings, consistent with pre-neural results, is beneficial.  
In particular, injecting discourse has been shown to either improve performance on the extractive summarization  task itself \cite{discourse-aware-extractive}, or allow for a substantial 
reduction in the number of the summarizer's parameters, while keeping competitive performance \cite{xiao-etal-2020-really}.

The central hypothesis we are exploring in this paper is 
whether the synergy between discourse parsing and summarization is bidirectional. In other words, we examine if 
summarization is 
a useful auxiliary task to infer discourse structures. \citet{liu-etal-2019-single} performed a preliminary investigation of this conjecture, showing that structural information can be inferred from 
attention mechanisms while training a neural model on auxiliary tasks. However, they did not perform any comparison against ground-truth discourse trees. Further, 
recent work showed that discourse trees implicitly induced during training are oftentimes trivial and shallow, not representing valid discourse structures \cite{ferracane2019evaluating}.




In this paper, we 
address these limitations by explicitly exploring the relationship between summarization and discourse parsing through the inference of document-level discourse trees from pre-trained summarization models, comparing the results against ground-truth RST discourse trees. 
Besides \citet{liu-etal-2019-single}, our idea and approach are inspired by recent works on extracting syntactic trees from pre-trained language models \cite{wu-etal-2020-perturbed} or machine translation approaches \cite{raganato-tiedemann-2018-analysis}, as well as previous work on knowledge graph construction from pre-trained language models \cite{knowledge_graph_from_lm}. 
Specifically, we generate full RST-style discourse trees from 
self-attention matrices of a pre-trained transformer-based summarization model. We use three different tree-aggregation approaches (CKY \cite{jurafsky2014speech}, Eisner \cite{eisner} and CLE \cite{Chuliu,edmond}), generating a set of constituency and 
dependency trees representing diverse discourse-related attributes. 

Our proposal is thereby addressing one of the key  
limitations in discourse parsing, namely the lack of large training corpora. 
We aim to overcome this limitation 
by generating 
a large number of reasonable quality discourse trees from a pre-trained summarization model, similar in spirit to what  \citet{huber-carenini-2020-mega} did with sentiment. 
Admittedly, 
the discourse information captured with our approach is  summarization task-specific, 
however, our generated discourse treebank can be combined with further task-dependent treebanks (e.g. from sentiment) to train more powerful discourse parsers in a multitask framework.


Generally speaking,  the ability to infer discourse trees as a ``by-product" of the summarization task can also be seen as a form of unsupervised discourse parsing, where instead of leveraging pre-trained language models like in \citet{kobayashi2019split}, we exploit a pre-trained neural summarizer. 


We empirically evaluate our method on three datasets with human RST-style annotations, covering different text genres. Multiple experiments show that the summarization model learns discourse information implicitly, and that more dependency information are captured, compared to structural (i.e., constituency) signals. Interestingly, an additional exploration of the attention matrices of individual heads suggests that, for all models, most of the discourse information is concentrated in a single head, and the best performing head is consistent across all datasets. We further find that the dependency information learned in the attention matrix covers long distance discourse dependencies. Overall, the results are 
consistent across datasets and models, 
indicating that the discourse information learned by the summarizer is general and transferable inter-domain.





 \vspace{-1mm}
\section{Related Work}
 \vspace{-1mm}

\textbf{Rhetorical Structure Theory (RST)} \cite{rst} is
one of the most popular theories of discourse, 
postulating that a document can be represented as a 
constituency tree, where leaves are clause-like Elementary Discourse Units (EDUs), 
and internal nodes combine their respective children by aggregating them into a single, joint constituent. Each internal node also has a nuclearity attribute\footnote{In this paper we do not consider rhetorical relations.}, representing the local importance of their direct child-nodes in the parent context from the set of \{Nucleus-Nucleus, Nucleus-Satellite, Satellite-Nucleus\}. ``Nucleus" child-nodes thereby generally play a more important role when compared to a ``Satellite" child-node.
Although standard RST discourse trees are encoded as constituency trees, they can be converted into dependency trees with near isomorphic transformations. In this work, we infer both, constituency and dependency trees.

Over the past decades, \textbf{RST discourse parsing} has been mainly focusing on supervised models, typically trained and tested within the same domain using human annotated 
discourse treebanks, such as RST-DT \cite{carlson2002rst}, Instruction-DT \cite{subba2009effective} or GUM  \cite{Zeldes2017}. The intra-domain performance of these supervised models has consistently improved, 
with a mix of 
traditional models by 
\citet{joty2015codra} and 
\citet{wang2017two}, and neural models 
\cite{yu2018transition} reaching state-of-the-art (SOTA) 
results. 
Yet, these approaches do not 
generalize well inter-domain \cite{huber-carenini-2020-mega}, likely due to the 
limited amount of available training data. 

\citet{huber2019predicting} recently tackled this data-sparsity issue through automatically generated discourse structures from distant supervision, showing that sentiment information can 
be used to infer discourse trees. Improving on their initial results, \citet{huber-carenini-2020-mega} published a large-scale, distantly supervised discourse corpus (MEGA-DT), showing that a parser trained on such treebank  delivers SOTA performance on the more general inter-domain discourse parsing task. In this paper, we also tackle the data sparsity problem in discourse parsing, however, using a significantly different approach. 
First, instead of relying on sentiment, we leverage the task of extractive summarization. Second, instead of a method for distant supervision, we propose an unsupervised approach.


The area of unsupervised RST-style discourse parsing has been mostly underlooked in the past, 
with recent neural approaches either  taking advantage of pre-trained language models to predict discourse \cite{kobayashi2019split} or using pre-trained syntactic parsers and linguistic knowledge \cite{nishida2020unsupervised} to infer discourse trees in an unsupervsied manner. 
Similarly. our proposal only relies on 
a pre-trained neural summarization model to generate discourse trees.

Recent \textbf{neural summarization models} are typically based on  transformers 
\cite{liu-lapata-2019-hierarchical, zhang-etal-2019-hibert}.
One advantage of these 
models is that they learn the relationship between input units explicitly using the dot-product self-attention, which allows for some degree of exploration of the inner working of these complex and distributed models.
Here, we  investigate 
if the attention matrices of a transformer-based
summarizer 
effectively capture discourse information (i.e., how strongly EDUs are related) and therefore can be used to derive discourse trees for arbitrary documents. 

\citet{Marcu1999} pioneered the idea 
to directly \textbf{apply RST-style
discourse parsing to 
extractive summarization}, 
and empirically showed that RST discourse information can 
benefit the summarization task, by simply extracting EDUs along the nucleus path. 
This initial success was followed by further work on leveraging discourse parsing in summarization, including \citet{knapsack_original}, \citet{TreeKnapsack}, and \citet{NestedTree}.  More recently, the benefits of discourse for summarization have also been confirmed for neural summarizers, e.g. in \citet{xiao-carenini-2019-extractive} and   \citet{discourse_abstractive}, using the 
structure of scientific papers (i.e. sections), and in \newcite{discourse-aware-extractive}, successfully incorporating RST-style
discourse and co-reference information in the BERTSUM summarizer \cite{bertsum}.

In contrast to previous approaches demonstrating how discourse 
can enhance summarization performance, we have recently shown that discourse enables the specification of simpler neural summarizers, without affecting their performance \cite{xiao-etal-2020-really}. In particular, by using a fixed discourse-based attention they achieve competitive results  compared to learnable dot-product self-attention mechanisms, as used in the original transformer model. 
Inspired by these findings, suggesting that transformer-based summarization models learn effective discourse representations, we explore if useful discourse structures can be inferred from learnt transformer self-attention weights.

Admittedly, \citet{liu_classification} and \citet{liu-etal-2019-single} presented preliminary work on  inferring discourse structures from attention mechanisms, while training a neural model on auxiliary tasks, like text classification and summarization. However, they did not perform any comparison against  ground-truth discourse trees as we do here. More importantly,  
we employ a more explicit approach to infer discourse structures, not as part of the learning process, but extracting the structures after the summarization model is completely trained and applied to new documents.


While our focus is on discourse, \textbf{extracting syntactic constituency and dependency trees} from transformer-based models has been recently attempted in both, machine translation and language modelling.
In machine translation, \citet{marecek-rosa-2019-balustrades} and \citet{ raganato-tiedemann-2018-analysis} show that trained 
translation models can capture syntactic information within their attention heads, using the CKY and CLE algorithms, respectively. 
In pre-trained language models, \citet{wu-etal-2020-perturbed} propose a parameter-free probing method to construct syntactic dependency trees 
based on a pre-trained BERT model, only briefly elaborating on possible applications to discourse. In contrast to our work, they  do not directly use attention heads, but instead build an impact matrix based on the distance between token representations. 
Furthermore, while their 
BERT-based model cannot deal with long sequences, our two-level encoder can effectively deal with sequences of any length, which is critical in discourse.


 \begin{figure*}[tb!]
    \centering
    \includegraphics[width=\linewidth]{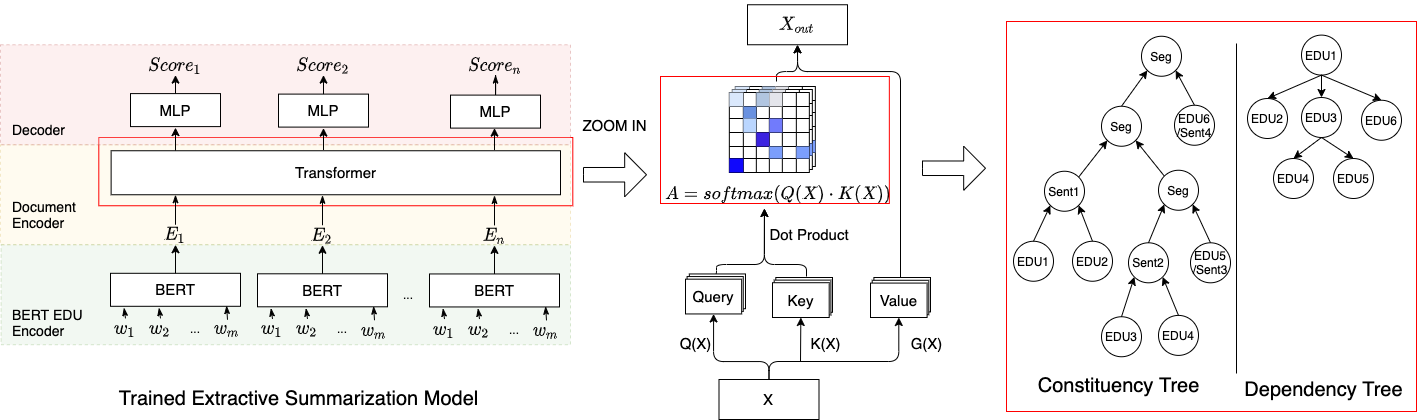}
    \caption{The pipeline of our whole method.}
    \label{fig:summarizer}
\end{figure*}

\section{Our Model}
 
 
 \vspace{-1mm}
\subsection{Framework 
Overview}
\label{overview}
\vspace{-1mm}
Our main goal 
is to show the ability of a previously trained summarization model to be directly applied to the task of RST-style discourse parsing. Along this line, we explore the relationship between information learned by the transformer-based sumarizer 
and the task of discourse parsing. 
We leverage the synergies between units learned in the transformer model 
by following 
\citet{xiao-etal-2020-really}, previously proposing the use of a transformer document-encoder on top of a pretrained BERT EDU
encoder.
This standard summarization model is presented in Figure \ref{fig:summarizer} (left). 
In the transformer-based document encoder, each head internally contains a self-attention matrix, learned during the training of the summarization model, representing the relationship between 
EDUs (Figure \ref{fig:summarizer} (center)). 
In this paper, we analyze these learned self-attention matrices, not only to confirm our intuition that they contain relevant discourse information, but also to computationally exploit such information for discourse parsing. 
We therefore generate a set of different (constituency/dependency) discourse trees from the self-attention matrices, focusing on different attributes of discourse, as shown in Figure \ref{fig:summarizer} (right).
Our generated constituency trees 
only reveal the discourse tree structure without additional nuclearity and relation attributes. 
More interestingly, we complement the constituency interpretation of the self-attention by additionally inferring a dependency tree, also partially guided by discourse structures, but mostly driven by the RST nuclearity attribute, which is shown to be more related to the summarization task where the importance of the different text spans is critical \cite{TreeKnapsack}. We present and discuss the different parsing algorithms to extract discourse information from the self-attention matrix next.
 \vspace{-1mm}
\subsection{Parsing Algorithms}
\label{parsing-algo}
\vspace{-1mm}
Formally, for
an input document $D=\{u_1, .. ,u_n\}$ with $n$ EDUs, each attention head returns an attention matrix $A\in \mathbf{R}^{n\times n}$ where entry $A_{ij}$ contains a score measuring how much the $i$-th EDU relies on the $j$-th EDU. Given those bidirectional scores defining the relationship between every two EDUs in a document, we build a tree
such that EDU pairs with higher reciprocal attention scores are more closely associated in the resulting tree. 
In the constituency case, this means that EDUs with higher mutual attention should 
belong to sub-trees on lower levels of the tree, 
while in the dependency case this implies that the path between such EDUs should contain less intermediate nodes. In essence, these requirements can be formalized as searching for the tree within the set of possible trees, 
which maximizes a combined score. 
 \vspace{-1mm}
\subsubsection{Constituency Tree (C-Tree) Parsing}
\label{ctree_parsing}
\vspace{-1mm}

To generate a constituency tree from the attention matrix, we follow a large body of previous work in discourse parsing (e.g., \citet{joty2015codra}), where constituency discourse trees are generated using the 
CKY algorithm \cite{jurafsky2014speech}.
Specifically, we 
fill a $n\times n$ matrix $P\in \mathbf{R}^{n\times n}$ generating the optimal tree in bottom-up fashion using the dynamic programming approach according to: 
\[
   P_{ij}= \begin{cases} 
        0, & i>j\\ 
        \sum_{k=1}^n(A_{ki}), & i=j \\
        \max_{k=i}^{j-1} ( P_{ik}+P_{(k+1)j}\\
        \quad +\avg(A_{i:k,(k+1):j})\\
        \quad +\avg(A_{(k+1):j,i:k}))/2, & i<j
        \end{cases}
\]
where $P_{ij}$ with $i=j$ contains the overall importance of EDU $i$, computed as the attention paid by others to unit $i$. $P_{ij}$ with $i < j$ represents the score of the optimal sub-tree spanning from EDU $i$ to EDU $j$. We select the best combination of sub-trees $k$, such that the sum of the left sub-tree spanning $[i:k]$  and the right one spanning $[(k+1):j]$, along with the average score of connections between the two sub-trees is maximized. 
\begin{figure}[h!]
    \centering
    \includegraphics[width=\linewidth]{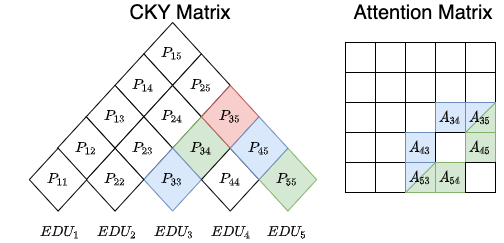}
    \caption{Example of CKY constituency parsing.}
    \label{fig:cky}
    \vspace{-3mm}
\end{figure}
\vspace{-1mm}
\begin{figure*}
    \centering
    \includegraphics[width=0.85\linewidth]{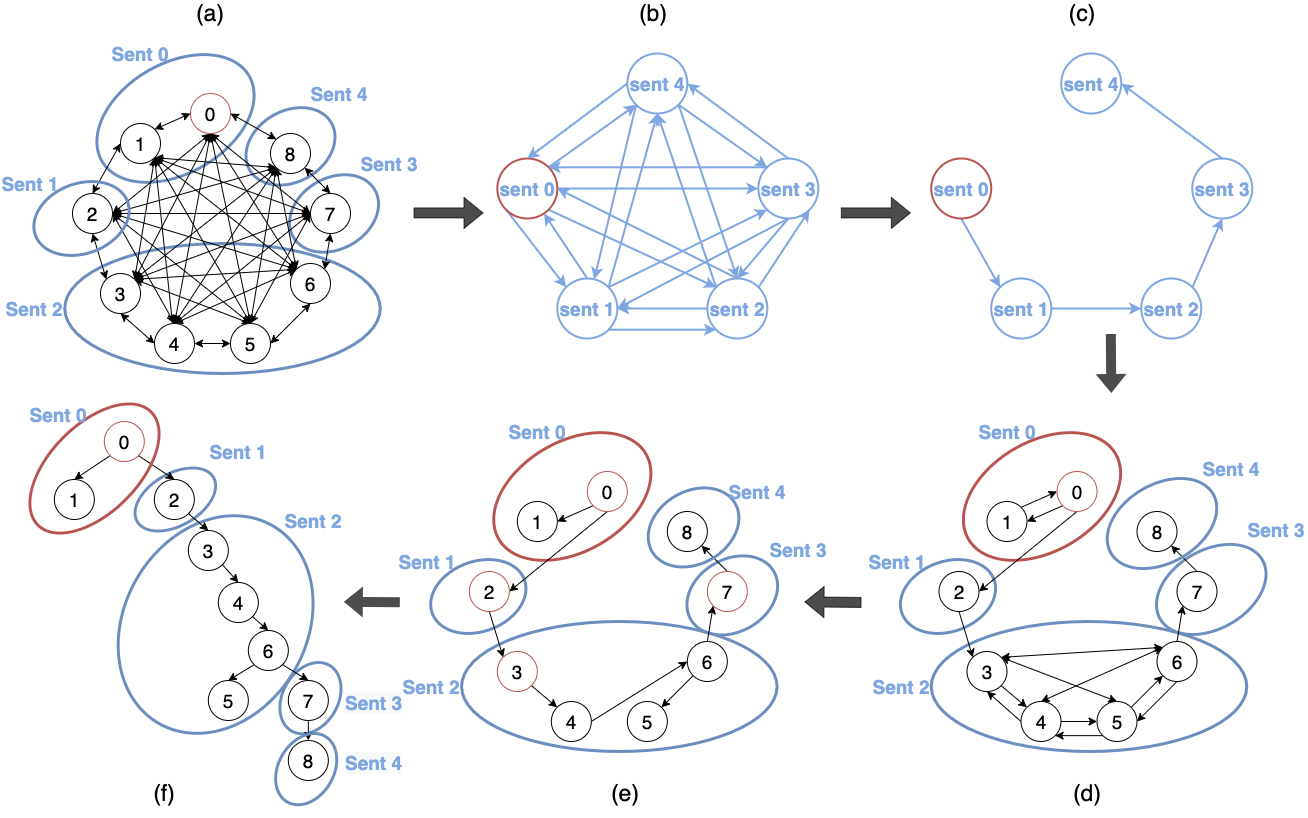}
    \caption{Chu-Liu-Edmonds Algorithm with sentence constraints}
    \label{fig:cle_sent}
    \vspace{-3mm}
\end{figure*}
For example, to pick the structure of the sub-tree spanning EDUs $[3:5]$ (
see Fig. \ref{fig:cky}), we need to decide between the potential sub-tree aggregation of $((34)5)$ and $(3(45))$. The respective scores are computed based on the scores in green and blue blocks in both the CKY and the Attention Matrices.
Following this algorithm, two sub-trees with a high attention score between them tend to be combined on lower levels of the tree, indicating they are more related in the discourse tree.

Besides the standard CKY algorithm described above, we also explore a hierarchical CKY approach with sentence and paragraph constraints. Specifically, we do not aggregate $P_{ij}$ if the span $[i:j]$ crosses a sentences boundary where either sentence is incomplete. In the previous example, 
if $EDU_3$ and $EDU_4$ 
were in the same sentence, even if the score of the blue aggregation candidate was higher, we would choose the green sub-tree aggregation. 
Plausibly, this hierarchical approach will perform better, since the ground-truth treebanks mostly contain sentences and paragraphs that are covered by a complete discourse sub-trees.

 \vspace{-1mm}
\subsubsection{Dependency Tree (D-Tree) Parsing}
For the dependency tree generation, we use the Eisner \cite{eisner} and Chu-Liu-Edmonds algorithm \cite{Chuliu,edmond} to generate projective and non-projective dependency trees, respectively
\footnote{``Mixed" approaches, dealing with mildly non-projective trees \cite{kuhlmann-nivre-2006-mildly}, are left for future work.}. 
First, we convert the attention matrix $A$ into a fully connected graph $G=(N,E)$, where $N$ contains all the EDUs, and $e_{ij}$, indicating how much the $i$-th EDU influences the $j$-th EDU, corresponds to $A_{ji}$, which is the attention that the $j$-th EDU pays to the $i$-th EDU. Based on this 
graph, we apply the following algorithms:
\vspace{-1mm}
\paragraph{Eisner Algorithm:}
We apply this dynamic programming algorithm to generate projective dependency trees. 
Thereby, we build a matrix $P\in \mathbf{R}^{n\times n\times 2 \times 2}$, in which the first and second dimensions contain the start and end indexes of sub-trees, similar to the CKY algorithm; while the third and fourth dimensions indicate whether the head is the start or the end unit, and whether the sub-tree is completed. 
As done for constituency parsing, we also use a hierarchical version of Eisner's algorithm, in which we restrict inter-sentence connections for incomplete sentence trees. 
Since the Eisner algorithm can only generate  projective  dependency trees, it will be inaccurate for documents with a non-projective discourse structure.

\vspace{-1mm}
\paragraph{Chu-Liu-Edmonds (CLE) Algorithm:}
Originally proposed as a recursive approach to find the maximum spanning tree of a graph given its root, CLE can generate non-projective trees.  
In the unconstrained case, we simply follow the standard CLE algorithm, selecting the EDU with the highest importance score, computed similar to Sec. \ref{ctree_parsing}, i.e. $root= \argmax_i \sum_{k=1}^n(A_{ki})$, as the root. From there, the algorithm selects the ``optimal edges", i.e. the maximum in-edges for each node except the root, breaking the cycles recursively. 

Again, as we did for CKY and Eisner, we also apply the additional sentence constraint. 
Unlike for the dynamic programming approaches, which build the trees in a bottom-up fashion and can directly be constrained to avoid cross-sentence aggregations of incomplete sentences, we need to substantially modify CLE to allow for sentence constraints. 

In particular, we first build a sentence graph $G^s=\{N^s,E^s\}$ from the EDU graph (Figure \ref{fig:cle_sent} (b)), in which $e^s_{SD}=\avg\limits_{s\in S,d\in D}e_{sd}$, and record the maximum edge corresponding to the edge between sentences, i.e. $\argmax_{s\in S,d\in D}e_{sd}$. After that,  we use the CLE algorithm within the sentence containing the root EDU as the root sentence to find the maximum spanning tree in $G^s$ (Figure \ref{fig:cle_sent} (c)). We then add the corresponding EDU edges to the final tree (Figure \ref{fig:cle_sent} (d)). For example, the edge $(s_0,s_1)$ in $G^s$ corresponds to the EDU edge $(e_0,e_2)$ in $G$. Next, we treat nodes with incoming edges from other sentences as the root of the sentence itself and run the CLE algorithm within each sentence (Figure \ref{fig:cle_sent} (e)). The final tree (Figure \ref{fig:cle_sent} (f)) is eventually formed as the combination of inter-sentence edges derived in sentence graph $G_s$ and intra-sentence edges found within each sentence.
 \vspace{-1mm}
\section{Experiments and Analysis}

\vspace{-1mm}
\subsection{The Summarization Task}
\label{summ_model}
\vspace{-1mm}
In order to show the generality of the discourse structures learned in the summarization model, we train our summarizer across a variety of datasets and hyper-parameter settings.
More specifically, 
we train on two separate, widely-used news corpora -- CNN Daily Mail (CNNDM) \cite{nallapati-etal-2016-abstractive} and NYT \cite{nytdata} --, as well as under three hyper-parameter settings with different numbers of layers and attention heads: 
(a) A simple model with 2 layers and a single head. 
(b) 6 layers with 8 heads each, proposed in the original transformer model\cite{transformer}.
(c) 2 layers with 8 heads each, constituting a middle ground between the previous two settings. By considering two corpora (CNNDM and NYT) and the three settings, we train six models, which we call: CNNDM-2-1, CNNDM-6-8, CNNDM-2-8, NYT-2-1, NYT-6-8, NYT-2-8\footnote{Complete evaluation results for all six models are presented in Appendix \ref{performance_summ}.}.


\vspace{-1mm}
\subsection{Discourse Datasets}
\label{datasets}
\vspace{-1mm}
The quality of the attention-generated trees is assessed 
on three 
discourse datasets (see Table \ref{tab:data_sta}). 

\textbf{RST-DT} is the largest and most frequently used RST-style discourse treebank \cite{carlson2002rst}, containing news articles from the Wall Street Journal. 
Since this is the genre of both our summarization training corpora, the experiments testing on this dataset are 
intra-domain.

\textbf{Instruction-DT} 
contains documents in the home-repair instructions domain \cite{subba2009effective}. We categorize the experiments on this dataset as 
cross-domain.

\textbf{GUM} 
contains documents from eight domains including news, interviews, academic papers and more \cite{Zeldes2017}. Since the GUM corpus is 
multi-domain, the 
performance on this dataset will reveal the generalizability of generated trees in a broader sense.

\begin{table}[h]
    \centering
    \resizebox{\linewidth}{!}{
    \begin{tabular}{c|c|c|c|c}
    Dataset & \# Docs& \#EDU/doc& \#Sent/doc& \#words/doc   \\
    \hline
      RST-DT   &385&56.6&22.5&549\\ 
      Instruction& 176&32.7&19.5&318\\
      GUM&127&107&45&874\\
      \hline
    \end{tabular}}
    \caption{Key RST-style discourse dataset dimensions.}
    \label{tab:data_sta}

\end{table}
\begin{table}[t]
    \centering
    \resizebox{0.9\linewidth}{!}{
    \begin{tabular}{c|r|r|r|r}
    \hline
    \multirow{2}{*}{Model}&\multicolumn{2}{c|}{No Cons.}&\multicolumn{2}{c}{Sent Cons.}\\
    \cline{2-5}
    &Attn\#0&Attn\#1& Attn\#0&Attn\#1\\
    \hline
    \multicolumn{5}{c}{RST-DT}\\
    \hline
    CNNDM-2-1&\color{OliveGreen}61.2 &\color{OliveGreen}59.7  &\color{OliveGreen}76.2 &\color{OliveGreen}74.6\\
    CNNDM-6-8&\color{OliveGreen}60.3 &\color{OliveGreen}60.8 &\color{OliveGreen}75.4 &\color{OliveGreen}75.0 \\
    NYT-6-8&\color{OliveGreen}\textbf{62.4} &\color{OliveGreen}62.2&\color{OliveGreen}\textbf{76.7} &\color{OliveGreen}75.6 \\
    \hline
    Random & \multicolumn{2}{c|}{58.6$\pm$0.1}&\multicolumn{2}{c}{74.1$\pm$0.1}\\
    \hline
    \multicolumn{5}{c}{Instruction}\\
    \hline
    CNNDM-2-1&\color{OliveGreen}61.1& \color{OliveGreen} 59.8&\color{OliveGreen}\textbf{71.4}&$\downarrow$ \color{Maroon}70.3\\
CNNDM-6-8& \color{OliveGreen} 60.3&\color{OliveGreen}\textbf{61.2}&\color{OliveGreen}71.2&\color{OliveGreen}70.9\\
NYT-6-8&\color{OliveGreen}\textbf{61.3}&\color{OliveGreen}61.3&\color{OliveGreen}71.3&$\downarrow$ \color{Maroon}70.0\\
    \hline
    Random & \multicolumn{2}{c|}{59.5$\pm$0.3}&\multicolumn{2}{c}{70.5$\pm$0.1}\\
    \hline
    \multicolumn{5}{c}{GUM}\\
    \hline
    CNNDM-2-1&\color{OliveGreen}58.7& \color{OliveGreen}57.7&\color{OliveGreen}\textbf{72.7}&\color{OliveGreen}71.9\\
CNNDM-6-8&\color{OliveGreen}58.9&\color{OliveGreen}59.3&\color{OliveGreen}72.4&\color{OliveGreen}\textbf{72.7}\\
NYT-6-8&\color{OliveGreen}\textbf{59.6}&\color{OliveGreen}59.3&\color{OliveGreen}72.2& \color{OliveGreen} 71.6\\

    \hline
    Random & \multicolumn{2}{c|}{57.5$\pm$0.1}&\multicolumn{2}{c}{71.5$\pm$0.2}\\
    \hline
    \end{tabular}}
    \caption{RST Parseval Scores of generated constituency trees on the three datasets, expressed as 'Avg. $\pm$ Std'.  \textcolor{OliveGreen}{Green} means the result is better than Random $+$ Std, and \textcolor{Maroon}{Red} along with $\downarrow$ means worse. Results for Random are obtained by applying the parser to random 
    matrices for 10 times. Attn\#0/1 are the first two layers.}
    \label{tab:const}
\vspace{-2mm}
\end{table}
All three discourse datasets 
contain ground-truth RST-style consituency trees. While all corpora contain potential non-binary sub-trees, Instruction-DT also includes multi-root documents. To account for these cases, we apply the right-branching binarization following \citet{huber2019predicting}. Furthermore, we convert constituency trees with nuclearity into ground truth dependency trees using the algorithm proposed in \citet{li-etal-2014-text} .

\vspace{-1mm}
\subsection{Evaluation Metric}
To evaluate how well the generated trees align with ground-truth trees, we use RST Parseval Scores 
for constituency trees and Unlabeled Attachment Score 
for dependency trees, measuring the ratio of matched spans and the ratio of matched dependency relations, respectively.
\subsection{Overall Results}
\label{results}
\vspace{-1mm}
For each model configuration, 
we run a set of 
experiments using the average attention matrix across all heads in a layer
, i.e. $A_{avg}=\sum_{h}A^h/H$, with $H$ as the number of heads
. This initial setup is intended to provide insights 
into the discourse information learned in each layer.

\begin{table}[]
    \centering
    \resizebox{0.9\linewidth}{!}{
    \begin{tabular}{c|r|r|r|r}
    \hline
    \multirow{2}{*}{Model}&\multicolumn{2}{c|}{No Cons.}&\multicolumn{2}{c}{Sent Cons.}\\
    \cline{2-5}
    &Attn\#0&Attn\#1& Attn\#0&Attn\#1\\
    \hline
    \multicolumn{5}{c}{RST-DT}\\
    \hline
CNNDM-2-1&\color{OliveGreen} \textbf{23.7}&$\downarrow$ \color{Maroon}4.8&\color{OliveGreen}\textbf{28.2}&$\downarrow$ \color{Maroon}18.2\\
CNNDM-6-8&$\downarrow$ \color{Maroon}7.9&\color{OliveGreen}20.5&$\downarrow$ \color{Maroon}13.8&\color{OliveGreen}27.8\\
NYT-6-8&\color{OliveGreen}15.7&\color{OliveGreen}12.5&\color{OliveGreen}24.3&$\downarrow$ \color{Maroon}18.9\\
    \hline
    Random & \multicolumn{2}{c|}{11.2$\pm$0.2}&\multicolumn{2}{c}{20.3$\pm$0.2}\\
    \hline
    \multicolumn{5}{c}{Instruction}\\
    \hline

CNNDM-2-1&\color{OliveGreen}\textbf{31.1}&$\downarrow$  \color{Maroon} 4.4&\color{OliveGreen}\textbf{29.3}&$\downarrow$ \color{Maroon}13.5\\
CNNDM-6-8&$\downarrow$ \color{Maroon}8.5&\color{OliveGreen}19.5&$\downarrow$ \color{Maroon}9.9&\color{OliveGreen}22.0\\
NYT-6-8&\color{OliveGreen}16.2&$\downarrow$ \color{Maroon}12.1&\color{OliveGreen}22.8&$\downarrow$ \color{Maroon}16.4\\

    \hline
    Random &
\multicolumn{2}{c|}{13.1$\pm$0.3}&\multicolumn{2}{c}{19.3$\pm$0.4}\\
    \hline
    \multicolumn{5}{c}{GUM}\\
    \hline
    
CNNDM-2-1&\color{OliveGreen}\textbf{21.3}&$\downarrow$ \color{Maroon}2.24&\color{OliveGreen}\textbf{27.3}&$\downarrow$ \color{Maroon}16.1\\
CNNDM-6-8&$\downarrow$ \color{Maroon}4.7&\color{OliveGreen}15.8&$\downarrow$ \color{Maroon}11.5&\color{OliveGreen}24.80\\
NYT-6-8&\color{OliveGreen}12.6&$\downarrow$ \color{Maroon}9.6&\color{OliveGreen}23.4&$\downarrow$ \color{Maroon}17.1\\

    \hline
    Random &
\multicolumn{2}{c|}{10.4$\pm$0.2}&\multicolumn{2}{c}{19.2$\pm$0.3}\\
    \hline
    \end{tabular}}
        \caption{ Unlabeled Attachment Scores of dependency trees generated by the Eisner algorithm.}
    \label{tab:eisner}

\end{table}
\begin{table}[h]
    \centering
    \resizebox{0.9\linewidth}{!}{
    \begin{tabular}{c|r|r|r|r}
    \hline
    \multirow{2}{*}{Model}&\multicolumn{2}{c|}{No Cons.}&\multicolumn{2}{c}{Sent Cons.}\\
    \cline{2-5}
    &Attn\#0&Attn\#1& Attn\#0&Attn\#1\\
    \hline
    \multicolumn{5}{c}{RST-DT}\\
    \hline
CNNDM-2-1&\color{OliveGreen}\textbf{21.6 } &$\downarrow$ \color{Maroon}1.5 &\color{OliveGreen}\textbf{29.3}&\color{OliveGreen}19.6\\
CNNDM-6-8&\color{OliveGreen}7.3&\color{OliveGreen}17.3&$\downarrow$ \color{Maroon}16.1 &\color{OliveGreen}28.5\\
NYT-6-8&\color{OliveGreen}13.7&\color{OliveGreen}10.6&\color{OliveGreen}25.0&\color{OliveGreen}21.1\\

    \hline
    Random & \multicolumn{2}{c|}{1.7$\pm$0.1}&\multicolumn{2}{c}{18.7$\pm$0.1}\\
    \hline
    \multicolumn{5}{c}{Instruction}\\
    \hline

CNNDM-2-1&\color{OliveGreen}\textbf{28.1}&$\downarrow$ \color{Maroon} 2.1&\color{OliveGreen}\textbf{37.4}&\color{OliveGreen}18.1\\
CNNDM-6-8&\color{OliveGreen}6.9&\color{OliveGreen}15.9&$\downarrow$ \color{Maroon}14.9&\color{OliveGreen}25.8\\
NYT-6-8&\color{OliveGreen}14.8&\color{OliveGreen}9.8&\color{OliveGreen}25.4&\color{OliveGreen}21.1\\

    \hline
    Random &
\multicolumn{2}{c|}{2.9$\pm$0.2}&\multicolumn{2}{c}{17.9$\pm$0.4}\\
    \hline
    \multicolumn{5}{c}{GUM}\\
    \hline
CNNDM-2-1&\color{OliveGreen}\textbf{19.5}&$\downarrow$ \color{Maroon}0.7&\color{OliveGreen}\textbf{28.8}&\color{OliveGreen}17.9\\
CNNDM-6-8&\color{OliveGreen}4.0&\color{OliveGreen}13.1&$\downarrow$ \color{Maroon}14.9&\color{OliveGreen}25.4\\
NYT-6-8&\color{OliveGreen}10.7&\color{OliveGreen}8.2&\color{OliveGreen}23.0&\color{OliveGreen}19.5\\
    \hline
    Random &
\multicolumn{2}{c|}{0.9$\pm$0.05}&\multicolumn{2}{c}{17.0$\pm$0.2}\\
    \hline
    \end{tabular}}
    \caption{Unlabeled Attachment Scores of dependency trees generated by the CLE algorithm}
    \label{tab:cle}
\vspace{-2mm}
\end{table}
The results of the three tree-generation algorithms
are shown in Table \ref{tab:const}, \ref{tab:eisner} and \ref{tab:cle} along with the performance of a random baseline obtained by running the algorithms on 10 random matrices.
Here, we present the results of three selected models, limited to the performance of the first two layers for the 6-layer models, to allow for a direct comparison to the 2-layer models\footnote{Results for all six models can be found in Appendix \ref{sec:full_results}.}.  
Across evaluations, the layer-wise performance 
within the same models are rather distinct, indicating that different 
properties are learned in the layers. 
This finding is in line with previous work \cite{contextualize}, especially given that 
the performance of each layer is consistent across constituency and dependency parsing outputs for all datasets. Furthermore, the more layers the summarization model contains, the smaller the performance gap between layers becomes. We believe that this could be caused by the 
discourse information being further spread across different layers. Generally, we observe that models trained on the CNNDM dataset perform better than models trained on the NYT corpus, despite the larger size of the NYT dataset. 
Plausibly, the superior performance of our models trained on CNNDM potentially reflects a higher diversity within documents in the CNNDM dataset.

Comparing the constituency tree performance in Table \ref{tab:const} against the dependency tree results in Tables \ref{tab:eisner} and \ref{tab:cle}, we can clearly see that the improvement of the constituency parsing approach over the random baseline is much smaller than the improvements for the generated dependency parse-trees. 
Presumably, this larger improvement for the dependency trees 
is 
due to the fact that dependency relationships (strongly encoding the nuclearity attribute) are more directly related to the summarization task than the plain structure information. This is in line with previous work on applying dependency trees to the summarization task \cite{TreeKnapsack, discourse-aware-extractive} and indicates that the learned attention matrices contain valid discourse information. 

\begin{figure*}
    \centering
    \includegraphics[width=0.8\linewidth]{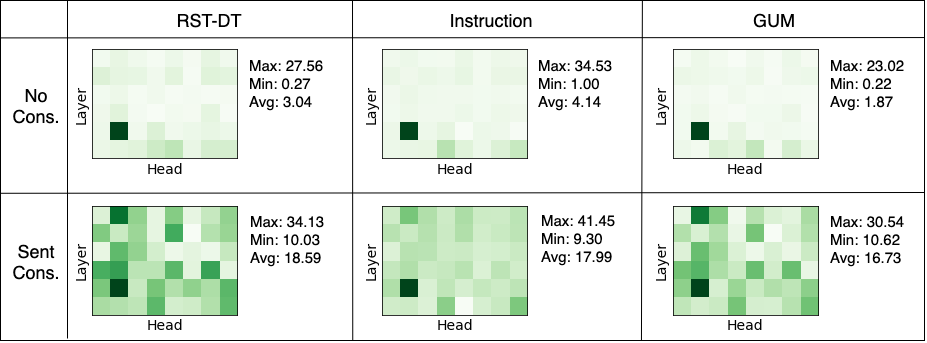}
    \caption{The Unlabeled Attachment Score of trees generated by the attention matrix per head on three datasets under two conditions with the \textit{CNNDM-6-8} model.}
    \label{fig:perhead}
    \vspace{-3mm}
\end{figure*}

As for the two approaches to dependency parsing, although Eisner generally outperforms CLE, 
the improvement over random trees is larger for CLE. 
We believe that this effect is due to the reduced constraints imposed on the CLE algorithm, which is not limited to generate projective trees. 

Considering all three methods, the results of the CLE-generated dependency tree seem most promising. 
A possible explanation is that both CKY and  Eisner 
 build the discourse tree in a bottom-up fashion with dynamic programming. This way, only local information is used on lower levels of the tree. On the other hand, the CLE algorithm uses global information, potentially more aligned with the summarization task, where all EDUs are considered to predict  importance scores. 
\vspace{-1mm}
\subsection{Performance of Heads}
\label{heads}
\vspace{-1mm}

While all previous results rely on the average attention matrices, we now analyze whether discourse information is evenly distributed across attention heads, or if a subset of the heads contains the majority of discourse related information.

We describe this analysis only for CLE for two reasons: (a) the summarization model seemingly captures more dependency-related discourse information than structure information; (b) compared with Eisner, 
the CLE approach is more flexible, by also covering non-projective dependency trees.


Since the results across all summarization models are consistent, we only show the accuracy heatmap for the \textit{CNNDM-6-8} model on the three RST-style discourse datasets 
in Figure \ref{fig:perhead}. 
Remarkably, for all three datasets, there is one head in the model capturing the vast majority of discourse information, especially in the unconstrained case. Furthermore, the performance of the best single attention head is much better than the one of the average attention matrix shown in section \ref{results} (e.g. $34.53$ compared to $19.51$ on the GUM dataset without sentence constraints). These intriguing findings will be further explored in future work.

\vspace{-1mm}
\subsection{Analysis of Generated Trees}
\vspace{-1mm}
\label{analysis}
\paragraph{Localness of Trees:} To further verify that the generated trees are non-trivial, for instance simply connecting adjacent EDUs, we 
analyze the quality of 
the trees produced with 
the second attention head on the second layer, which is the top performer among all the heads shown in Figure \ref{fig:perhead}. 
First, we separate all dependency relationships into 
two classes: {\it local}, 
holding between two adjacent EDUs, 
and {\it distant}, 
including all other relations between non-adjacent EDUs.
\begin{table}[]
    \centering
    \resizebox{0.75\linewidth}{!}{
    \begin{tabular}{c|c|c}
   Measurement(\%) & No Cons. & Sent Cons. \\
    \hline
    \multicolumn{3}{c}{RST-DT}\\
    \hline
    Local Ratio Corr. &77.78&79.17\\
    Local Ratio GT &\multicolumn{2}{c}{53.22}\\
    Local Ratio Ours &46.52&58.35\\
    \hline
    \multicolumn{3}{c}{Instruction}\\
    \hline
   Local Ratio Corr. &81.15&84.90\\
    Local Ratio GT &\multicolumn{2}{c}{59.82}\\
    Local Ratio Ours &47.90&60.54\\
    \hline
    \multicolumn{3}{c}{GUM}\\
    \hline
    Local Ratio Corr. &77.99&80.20\\
    Local Ratio GT & \multicolumn{2}{c}{53.28}\\
   Local Ratio Ours &39.97&53.76\\
    \hline
    \end{tabular}}
    \caption{Measurements on the locality of the generated dependency trees, all numbers are 
    in $\%$. Corr. represents all the correct predictions, GT the ground-truth trees, and Ours the generated tree respectively.}
    \label{tab:localness}
    \vspace{-4mm}
\end{table}
Then we compute the ratio of the correctly predicted dependencies which are local 
(Local Ratio Corr.), as well as the ratio of local dependencies in the generated trees (Local Ratio Ours), and in the ground-truth trees (Local Ratio GT). The results of this analysis are shown in Table \ref{tab:localness}. For all 
datasets, the ratio of correctly predicted local dependencies (Local Ratio Corr.) (being $>50$) 
is larger than the ratio for distant relations, 
which appears reasonable, since local dependency predictions are easier to predict than distant ones. Further, comparing 
(Local Ratio GT) and 
(Local Ratio Ours)
without the sentence constraint (first column) shows that the number of local dependency relations in the ground-truth discourse trees is consistently larger than the predicted number. 
This indicates that the discourse information learned in the attention matrices goes beyond the oftentimes predominant local positional information. However, even without the sentence constraint (first column), when the CLE algorithm can predict trees of any form, 
more than $40\%$ of the relations are predicted as local, suggesting that the standard CLE approach can already capture local information well.

Adding the sentence constraint (second column), we find that the local dependency ratio of the generated trees  
(Local Ratio Ours) further increases by more than $10\%$ across all three datasets. This makes intuitive sense, since the sentence constraint forces the generated trees to purely focus on local aspects within each sentence. 
To sum up, we find that the learned attention matrices contains both local and distant dependency information, although local dependency predictions perform better.
\paragraph{Properties of Trees:} Following \citet{ferracane2019evaluating}, we 
structurally inspect the generated dependency trees, and compare them with the gold 
trees on all three datasets. 
This 
comparison 
is presented in Table \ref{tab:properties}, showing the \textit{average branch width}, \textit{average height}, \textit{average leaf ratio (micro)} and \textit{average normalized arc length} of the trees as well as the \textit{percentage of vacuous trees} in each dataset\footnote{A vacuous tree is a special tree in which the root is one of the first two EDUs, with  
all nodes are children of the root.
}.

Looking at Table \ref{tab:properties}, 
it appears that our tree structure properties are similar to the ground-truth properties in regards to all measures except the height of the tree, which indicates that our trees tend to be generally deeper than gold standard trees, despite having a 
similar branch width and leaf ratio. 
Furthermore, 
our trees are even deeper when using the sentence constraint. Plausibly, by forcing each sentence to have its own sub-tree can make shallower inter-sentential structures less likely. Exploring potential causes for the difference in tree-height, possibly due to the summarization task itself, are left as future work.
\begin{table}[]
    \centering
    \resizebox{\linewidth}{!}{
    \begin{tabular}{c|c|c|c|c|c}
         &  Branch & Height & Leaf  & Arc & vac. (\%) \\
         \hline
        \multicolumn{6}{c}{RST-DT}\\
        \hline
        Ours(Sent Cons)&1.50&27.06&0.37&0.10&3\% \\
        Ours(No Cons)& 1.74 & 25.76 & 0.49&0.12&3\% \\
        GT Tree&2.10&8.19&0.51&0.13&2\% \\
         \hline
        \multicolumn{6}{c}{Instruction}\\
        \hline
        Ours(Sent Cons)&1.56&15.74&0.39&0.13&3\% \\
        Ours(No Cons)&1.80&14.35&0.50&0.14&3\% \\
        GT Tree& 1.59&8.49&0.41&0.15&1\% \\
         \hline
        \multicolumn{6}{c}{GUM}\\
        \hline
        Ours(Sent Cons)&1.61&44.94&0.40&0.05&0\% \\
        Ours(No Cons)& 2.14&43.08&0.54&0.08&0\% \\
        GT Tree& 2.02 & 12.17&0.51&0.04&0\% \\
        \hline
    \end{tabular}}
    \caption{Statistics of our generated trees and the gold standard trees in terms of the \textit{average branch width}, \textit{average height}, \textit{average leaf ratio (micro)}, \textit{average normalized arc length} of the trees and \textit{percentage of the Vacuous trees}.}
    \label{tab:properties}
    \vspace{-3mm}
\end{table}

\subsection{Additional Results on Model Sensitivity to Initialization and Summarizer Quality}
To investigate whether the performance is consistent cross different random initializations, and 
to explore the influence of the results with respect to the quality of the summarizer, we perform additional experiments with the 'CNNDM-6-8' model\footnote{More details can be found in Appendix \ref{sec:results_random_ablation}.}. 
Overall, we find that 
the performance is rather similar 
across random initializations. Interestingly, 
a single head consistently shows 
better performance than all other heads across 
different initialization as well as 
datasets; however, 
while the position of the top-performing head is not always the same, 
it is often 
located in the second layer of the model. 
Regarding the second experiment exploring sensitivity to the summarizer quality, 
we create summarizers of increasing quality by providing more and more  training. 
As expected, we find that as the summarization model is trained for additional steps, more accurate discourse information is learnt, concentrated in a single head.


\section{Conclusions and Future Work}
\vspace{-1mm}
We present a novel framework to infer discourse trees from the attention matrices learned in a transformer-based summarization model. Experiment across models and datsets indicates
that both dependency and structural discourse information are learned, that  such information is typically concentrated in a single head, and that the attention matrix also covers long distance discourse dependencies. 
Overall, consistent results across datasets and models suggest that the learned discourse information is general and transferable inter-domain. 

In the future, we 
want to explore if simpler summarizers 
like BERTSUM \cite{bertsum} can also capture discourse info; 
specifically studying if the importance of the heads corresponds to the captured discourse info, 
which may help pruning summarization model by incorporating discourse info, in spirit of \citet{xiao-etal-2020-really}.

With respect to dependency tree generation possible improvements 
could come by  looking for additional strategies 
balancing between guidance and flexibility,
as \citet{kuhlmann-nivre-2006-mildly} explore for syntactic dependency parsing.

To address the problem of data sparsity in discourse parsing, we want to  synergistically leverage other discourse-related tasks, in addition to sentiment and summarization, like topic modeling.




\vspace{-2mm}
\section*{Acknowledgments}
\vspace{-2mm}
We thank the anonymous reviewers and the UBC-NLP group for their insightful comments. This research was supported by the Language \& Speech Innovation Lab of Cloud BU, Huawei Technologies Co., Ltd.\\
We further acknowledge the support of the Natural Sciences and Engineering Research Council of Canada (NSERC).\\
Nous remercions le Conseil de recherches en sciences naturelles et en génie du Canada (CRSNG) de son soutien.
\bibliography{anthology,custom}
\bibliographystyle{acl_natbib}


\label{sec:appendix}
\appendix
\section{Performance of the Summarization Task}
\label{performance_summ}
Table.\ref{tab:permance_summarizer} shows the performance of different summarization models. In general, adding additional layers and heads does not consistently increase the performance on the summarization task itself.
\begin{table}[h]
    \centering
    \resizebox{\linewidth}{!}{
    \begin{tabular}{c|c|c|c|c|c}
    Dataset & \#Layer & \#Head & R-1 &R-2 & R-L\\
    \hline
    CNNDM&2&1&40.92&18.69&37.85\\
    CNNDM&2&8&41.02&18.78&37.96\\
    CNNDM&6&8&41.03&18.69&37.86\\
    \hline
    NYT&2&1&43.64&25.58&36.87\\
    NYT&2&8&44.11&26.08&37.34\\
    NYT&6&8&43.93&25.99&37.15\\
    \end{tabular}}
    \caption{The in-domain performance of the summarizers.}
    \label{tab:permance_summarizer}
\end{table}
\begin{table*}[]
    \centering
    \resizebox{\linewidth}{!}{
    \begin{tabular}{c|c|c|c|c|c|c}
    \hline
    \multirow{2}{*}{Model}&\multicolumn{2}{c|}{Const}&\multicolumn{2}{c|}{Eisner}&\multicolumn{2}{c}{CLE}\\
    &\multicolumn{1}{c|}{No Cons.}&\multicolumn{1}{c|}{Sent Cons.}&No Cons. &Sent Cons.&No Cons. &Sent Cons.\\
    \hline
   & \multicolumn{6}{c}{RSTDT}\\
    \hline
    CNNDM-2-1&61.16  /  59.67&76.23 / 74.63&\textbf{23.65} / 4.80&\textbf{28.24} / 18.23&\textbf{21.56} / 1.45&\textbf{29.29} / 19.56\\
    CNNDM-2-8&62.65 / 59.75&76.42 / 74.28&22.09 / 8.40&26.23 / 21.29&20.31 / 6.13&26.57 / 22.67\\
    CNNDM-6-8&60.33 / 60.79&75.44 / 75.04&7.89 / 20.48&13.83 / 27.78&7.28 / 17.30&16.10 / 28.50\\
    NYT-2-1&60.27 / 60.23&75.57 / 75.29&9.76 / 14.84&23.18 / 20.61&6.18 / 12.68&21.06 / 21.73\\
    NYT-2-8&\textbf{63.20} / 59.65&76.63 / 75.23&7.35 / 9.74&16.04 / 21.27&6.44 / 7.09&16.72 / 22.90\\
    NYT-6-8&62.42 / 62.17&\textbf{76.65} / 75.58&15.74 / 12.51&24.30 / 18.90&13.71 / 10.59&25.04 / 21.14\\
    \hline
    Random &58.60 (0.1)&74.10 (0.1)& 11.16 (0.2)&20.28 (0.2)& 1.67 (0.08)&18.72 (0.11)\\
    \hline
    &\multicolumn{6}{c}{Instruction}\\
    \hline
CNNDM-2-1&61.06 / 59.84&71.39 / 70.29&\textbf{31.07 }/ 4.39&\textbf{29.33} / 13.45&\textbf{28.06} / 2.08&\textbf{37.38 }/ 18.12\\
CNNDM-2-8&61.44 / 60.55&71.13 / 71.09&26.98 / 8.89&24.72 / 14.75&24.56 / 5.23&28.70 / 20.58\\
CNNDM-6-8&60.32 / 61.22&71.24 / 70.88&8.53 / 19.51&9.93 / 21.96&6.92 / 15.85&14.93 / 25.78\\
NYT-2-1&60.31 / 61.30&71.40 / \textbf{71.43}&10.67 / 21.15&18.99 / 21.19&7.82 / 17.59&21.30 / 24.54\\
NYT-2-8&61.27 / 60.51&70.80 / 70.90&6.26 / 12.59&13.64 / 19.34&5.25 / 7.96&13.39 / 21.92\\
NYT-6-8&\textbf{61.32} / 61.27&71.30 / 70.03&16.22 / 12.14&22.79 / 16.37&14.81 / 9.81&25.44 / 21.10\\
    \hline
    Random & 59.49 (0.3)&70.53 (0.1)&13.14 (0.33)&19.31 (0.44)&2.94 (0.24)&17.88 (0.42)\\
    \hline
    & \multicolumn{6}{c}{GUM}\\
    \hline
CNNDM-2-1&58.74 / 57.69&\textbf{72.73} / 71.92&\textbf{21.28} / 2.24&\textbf{27.26} / 16.12&\textbf{19.50} / 0.70&\textbf{28.77} / 17.92\\
CNNDM-2-8&59.98 / 58.43&72.69 / 71.95&19.45 / 4.98&25.00 / 19.25&18.03 / 2.92&25.07 / 20.40\\
CNNDM-6-8&58.92 / 59.30&72.40 / 72.69&4.74 / 15.80&11.53 / 24.79&4.01 / 13.14&14.85 / 25.37\\
NYT-2-1&57.81 / 58.84&71.95 / 72.23&5.64 / 12.84&19.94 / 20.19&2.92 / 9.79&18.23 / 19.68\\
NYT-2-8&\textbf{60.17} / 58.22&71.98 / 71.82&5.66 / 7.22&15.21 / 18.81&4.54 / 3.96&15.25 / 19.31\\
NYT-6-8&59.62 / 59.25&72.19 / 71.56&12.58 / 9.61&23.35 / 17.14&10.67 / 8.23&22.99 / 19.53\\

    \hline
    Random & 57.47 (0.1)&71.50 (0.2)&10.37 (0.23)&19.15 (0.26)&
0.92 (0.05)&17.01 (0.2)\\
    \hline
    \end{tabular}}
    \caption{The RST Parseval Scores of generated constituency trees, Unlabeled Attachment Score of generated dependency trees by Eisner algorithm and CLE algorithm on the three datasets. The numbers in each cell are represented as the performance of (Layer 0 / Layer 1) the results of Random are obtained by applying the parser on random generated matrices for 10 times, and are represented as 'Average (Std)'.}
    \label{tab:fulltable}
\end{table*}
\section{Full Results on Overall Tree Parsing}
\label{sec:full_results}
We show the overall results of all the six summarization models on constituency/dependency parsing in Table.\ref{tab:fulltable}, the results of three of them are shown in Table.\ref{tab:const}, Table.\ref{tab:eisner} and Table.\ref{tab:cle} in the main paper.

\begin{table*}[]
    \centering
    \resizebox{\linewidth}{!}{
    \begin{tabular}{c|c|c|c|c|c|c}
    \hline
    \multirow{2}{*}{Model}&\multicolumn{2}{c|}{Const}&\multicolumn{2}{c|}{Eisner}&\multicolumn{2}{c}{CLE}\\
    &\multicolumn{1}{c|}{No Cons.}&\multicolumn{1}{c|}{Sent Cons.}&No Cons. &Sent Cons.&No Cons. &Sent Cons.\\
    \hline
    &\multicolumn{6}{c}{RSTDT}\\
    \hline
\multirow{2}{*}{CNNDM-6-8}&61.13 / 61.63&75.81 / 75.41&10.32 / 20.99&16.42 / 27.08 &9.40 / 18.16&18.89 / 28.33\\
&(1.11) / (1.35)&(0.26) / (0.34)&(4.03) / (2.80)&(3.62) / (1.37)&(3.92) / (3.25)&(4.19) / (1.59)\\
    \hline
    Random & 58.6 (0.1)&74.10 (0.1)& 11.16 (0.2)&20.28 (0.2) & 1.67 (0.08)&18.72 (0.11)\\
    \hline
    &\multicolumn{6}{c}{Instruction}\\
    \hline
\multirow{2}{*}{CNNDM-6-8}&61.87 / 61.06&70.84 / 70.94&11.50 / 19.78&12.91 / 22.45&9.79 / 16.53&17.81 / 26.30\\
&(1.17) / (0.98)&(0.51) / (0.34)&(5.71) / (2.17)&(4.39) / (1.48)&(5.31) / (2.73)&(4.70) / (1.64)\\
    \hline

    Random & 59.49 (0.3)&70.53 (0.1)&13.14 (0.33)&19.31 (0.44)&2.94 (0.24)&17.88 (0.42)\\
    \hline
    &\multicolumn{6}{c}{GUM}\\
    \hline
\multirow{2}{*}{CNNDM-6-8}&58.19 / 58.71&72.28 / 72.48&7.62 / 15.69&14.62 / 24.02&6.77 / 13.23&17.32 / 25.13\\
&(0.82) / (0.97)&(0.27) / (0.16)&(3.97) / (2.87)&(3.77) / (1.36)&(3.96) / (3.29)&(3.85) / (1.22)\\
    \hline
    Random & 57.47 (0.1)&71.50 (0.2)&10.37 (0.23)&19.15 (0.26)&
0.92 (0.05)&17.01 (0.2)\\
    \hline
    \end{tabular}}
    \caption{The average RST Parseval Scores of generated constituency trees, the average Unlabeled Attachment Scores of generated dependency trees by the Eisner and CLE algorithms, on the three datasets with 5 random initialization, the numbers in parenthesis are the standard deviation across different run}
    \label{tab:rand-init_full}
\end{table*}

\begin{figure*}
    \centering
    \includegraphics[width=0.7\linewidth]{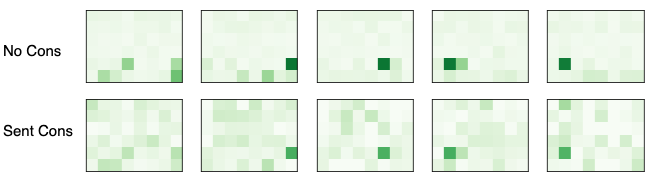}
    \caption{The heatmap of average UAS across three discourse datasets for all attention heads in the models with different initialization by the CLE algorithm.}
    \label{fig:random_init_heatmap}
\end{figure*}

\section{Results on Sensitivity to Initialization and Summarizer Quality}
\label{sec:results_random_ablation}
To explore if the models with different random initialization have consistent performances, we train 5 models with 6 layers and 8 heads on the CNNDM dataset with different initialization, and the results of each layer for constituency/dependency parsing are shown in Table.\ref{tab:rand-init_full}.
We can find that the results are relatively consistent across different initialization, and additional exploration on the performance of all the heads (Fig.\ref{fig:random_init_heatmap}) show that, with different initialization of the model, there is consistently one head containing most of the discourse information, but the position of that head is not fixed.

\begin{figure*}
    \centering
    \includegraphics[width=0.8\linewidth]{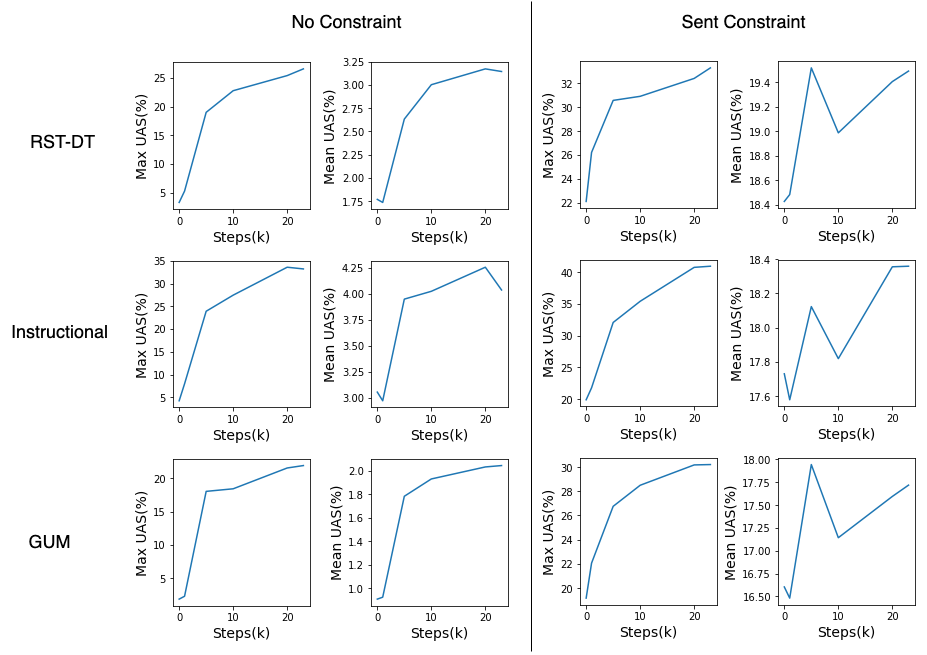}
    \caption{Max and Mean UAS of dependency trees generated by CLE algorithm on all attention heads (48) of the model 'cnndm-6-8', after training for (0,1,5,10,20,23) K steps on RST-DT(top), Instructional(middle) and GUM(bottom) datasets. The corresponding ROUGE scores are increasing.}
    \label{fig:error_pg_all}
\end{figure*}

\begin{figure*}
    \centering
    \includegraphics[width=0.85\linewidth]{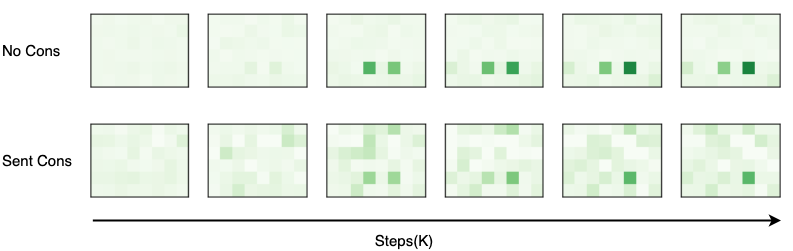}
    \caption{The heatmaps of the average UAS across the three discourse datasets for all the heads during training the summarization model.}
    \label{fig:heatmap_error_pg}
\end{figure*}

We further do the experiments on dependency parsing during training the summarizer, to see how the performance changes as the summarizer become better, and show the max and mean UAS over three datasets for all attention heads in the 'cnndm-6-8' model by the CLE algorithms after training for (0, 1k, 5k, 10k,  20k) steps in Fig.\ref{fig:error_pg_all}. We also show the heatmaps of the average UAS across three datasets for all the heads in Fig.\ref{fig:heatmap_error_pg}. We can find that as the summarizer is trained for more steps, more discourse information is learned, and it's more concentrated in one head. Interestingly, the mean UAS of dependency trees generated by CLE algorithm with sentence constraints show a different trend, which may due to the concentration of the discourse information on single head as the model trained for more steps, as it shows in Figure.\ref{fig:heatmap_error_pg}.



\end{document}